\documentclass[12pt, onecolumn, conference]{IEEEtran}
\IEEEoverridecommandlockouts
\usepackage{cite}
\usepackage{amsmath,amssymb,amsfonts}
\usepackage{algorithm}%
\usepackage{algorithmicx}%
\usepackage{algpseudocode}%
\usepackage{graphicx}
\usepackage{booktabs}
\usepackage{tabularx}
\usepackage{multirow, multicol, array}
\usepackage{wrapfig}
\usepackage{textcomp}
\usepackage{xcolor}
\usepackage{siunitx}
\renewcommand{\baselinestretch}{1.5}
\def\BibTeX{{\rm B\kern-.05em{\sc i\kern-.025em b}\kern-.08em
    T\kern-.1667em\lower.7ex\hbox{E}\kern-.125emX}}

\usepackage{todonotes}

\newcolumntype{L}[1]{>{\raggedright\arraybackslash}p{#1}}
\newcolumntype{C}[1]{>{\centering\arraybackslash}p{#1}}
\newcolumntype{R}[1]{>{\raggedleft\arraybackslash}p{#1}}

\newcolumntype{Y}{>{\centering\arraybackslash}X}
\newcolumntype{Z}[1]{>{\raggedright\arraybackslash}p{#1}}

\begin{document}


\title{Convolutional Neural Networks Exploiting Attributes of Biological Neurons}

\author{\IEEEauthorblockN{Neeraj Kumar Singh}
\IEEEauthorblockA{\textit{Computer Vision and Pattern Recognition Unit} \\
\textit{Indian Statistical Institute}\\
Kolkata, India \\
neeraj1909@gmail.com}
\and
\IEEEauthorblockN{Nikhil R. Pal}
\IEEEauthorblockA{\textit{Electronics and Communication Sciences Unit} \\
\textit{Indian Statistical Institute}\\
Kolkata, India \\
nrpal59@gmail.com}
}

\maketitle

\begin{abstract}
In this era of artificial intelligence, deep neural networks like Convolutional Neural Networks (CNNs) have emerged as front-runners, often surpassing human capabilities. These deep networks are often perceived as the panacea for all challenges. Unfortunately, a common downside of these networks is their ''black-box'' character, which does not necessarily mirror the operation of biological neural systems. Some even have millions/billions of learnable (tunable) parameters, and their training demands extensive data and time.

Here, we integrate the principles of biological neurons in certain layer(s) of CNNs. Specifically, we explore the use of neuro-science-inspired computational models of the Lateral Geniculate Nucleus (LGN) and simple cells of the primary visual cortex. By leveraging such models, we aim to extract image features to use as input to CNNs, hoping to enhance training efficiency and achieve better accuracy.  We aspire to enable shallow networks with a Push-Pull Combination of Receptive Fields (PP-CORF) model of simple cells as the foundation layer of CNNs to enhance their learning process and performance. To achieve this, we propose a two-tower CNN, one shallow tower and the other as ResNet 18.  Rather than extracting the features blindly, it seeks to mimic how the brain perceives and extracts features. The proposed system exhibits a noticeable improvement in the performance (on an average of $5\%-10\%$) on CIFAR-10, CIFAR-100, and ImageNet-100 datasets compared to ResNet-18. We also check the efficiency of only the Push-Pull tower of the network.
\end{abstract}


\section{Introduction}
\subsection{Deep Learning}
The biological brain serves as an inspiration for artificial neural networks (ANNs). ANNs typically use many neurons organized in layers resembling those we find in living things. Learned information is passed on through the link between these neurons. The recipient neurons then process and combine these patterns. Each link has a  weight denoting its relevance. For example, if a neuron has $m$ inputs, the weights of these $m$ links determine the significance of the received information. Furthermore, it is worth mentioning that a link's weight can either amplify or suppress the signal.

Deep learning encompasses a subset of machine learning and emphasizes the representation of the learning powers of ANNs. The adjective ``deep" in deep learning denotes the network's multiple (usually two or many more) hidden layers. Convolutional neural networks (CNNs) excel at capturing intricate decision boundaries. It includes convolutional layers designed to adaptively learn spatial hierarchical features, making them efficient for many tasks like image recognition, video recognition, and medical image analysis, to name a few.

Each layer in deep learning extracts specific patterns and passes them to the subsequent layer. Layer-by-layer training~\cite{kulkarni2017layer} and end-to-end training~\cite{mou2015end}, both can be used to create deep learning architectures. Many deep learning frameworks include pre-built methods, classes, and functions for constructing these systems.

Several elements have attributed to the growing interest in deep learning:
\begin{itemize}
    \item Availability of the extensive sets of labelled data: The effectiveness of deep learning models often increases with data volume. However, once the dataset reaches a particular size (assessing that the training dataset has sufficient instances generated from a time-invariant distribution), the performance of conventional networks typically reaches a plateau.    
    \item The shift to GPU-based parallel computing: This enabled a transition from CPU-driven training to GPU-driven training, significantly accelerating the training of deep learning models. Today, GPUs are required for training every deep-learning model.
    \item Superior feature extraction by deep models: Deep models have demonstrated to be extraordinarily good at extracting relevant features from data, allowing us to tackle complex decision-making tasks that were not feasible with conventional neural networks.
\end{itemize}

\subsection{Some Issues with Deep Learning}
When trained on large data sets, deep learning algorithms often excel over other learning systems but have their own hurdles. It has shown remarkable performance in computer vision and natural language processing. Yet, understanding the semantics of the feature maps learned in the hidden layers remains a mystery, limiting their use in high-stake domains like healthcare and defence. Another drawback of deep neural networks is their inability to explain their decisions clearly. This challenge becomes more pronounced when such a system possesses billions of free parameters. Imagine a deep network that classifies images of dogs and cats, which might result in impressive accuracy. However, the rationale behind its performance is unclear. We might identify the nodes(neurons) triggered by the input instance, yet comprehending the significance of these activated neurons remains difficult/ambiguous. It is unknown how the feature maps from each layer are integrated step by step to create the final feature map, what these neurons accomplish together, and how these extracted feature maps are merged to create the final feature map. Consequently, the inner workings of these deep networks remain like a black box.

However, DNNs (deep neural networks) are frequently considered "all-cure" remedies. These deep networks typically do not resemble how exactly biological neural networks work. Another significant problem is that, when faced with input from an unfamiliar class, most of these networks cannot respond with "Don't know." This restriction lessens the trustworthiness of such networks since it assumes a closed-world scenario, but, in reality, most classification problems involve an "open world" situation~\cite{chakraborty2007strict, karmakar2018make}. 

So, a few questions arise. Can we enhance the model's performance by leveraging our understanding of the brain? Could this lead to faster optimization of the model's parameters? Even if the answers are positive, it might not alter the black-box characteristic of CNNs. However, using the computational models inspired by the brain's information processing to improve the model's performance could be a step in the right direction. 

In our research, we aim to integrate the operations of biological neurons into specific layer(s) of CNNs. Specifically, we explored the Push-Pull CORF (combination of receptive fields) computational framework associated with Lateral Geniculate Nucleus (LGN) cells, as proposed in reference~\cite{azzopardi2012corf, azzopardi2014push}. These models derive features from images, subsequently fed into a CNN. We expect this approach to considerably enhance the model's efficacy, as it does not just extract features arbitrarily but by mimicking the brain's feature extraction process to a certain degree.




\section{Related Work}
Over the past several years, there have been many attempts to develop computational models of the functionalities of biological neurons, such as Ganglion cells and LGN cells. It has been found that the current deep learning architectures do not exactly operate on the principle of biological neurons. Yet, these models are inspired by how the mammal's brain processes the information passed by the retina. Inspired by the processing of these biological neurons, several computational models have been developed. Azzopardi et al.~\cite{azzopardi2013automatic} employ a collection of learnable keypoint detectors called Combination Of Shifted Filter Responses (COSFIRE) filters for the automatic identification of vascular junctions in the segmented retinal images. This method draws inspiration from a particular type of shape-distinguishing neuron located in the V4 region of the visual cortex. They set up COSFIRE filters attuned to various prototype bifurcations, showing that these filters are proficient at spotting bifurcations resembling the prototype examples. In another experiment~\cite{azzopardi2016increased}, Azzopardi et al. proposed an optimization technique based on the learnable COSFIRE filters. This method identifies the best combination of the contour segments for the desired pattern. It outperforms the original COSFIRE approach with substantially higher generalization capability, primarily in recall rate. Azzopardi et al. extended their work on COSFIRE filters and introduced a new method for identifying gender using the adaptable COSFIRE filters~\cite{azzopardi2016gender}. The selectivity of the COSFIRE filter is determined by analyzing a specific prototype pattern. They demonstrated the generalization ability of their method on GENDER-FERET and Labeled Faces in the Wild (LFW) datasets. Their approach outperforms the two commercial systems, namely Face++ and Luxand.
Extending the work on COSFIRE filters, Gecer et al.~\cite{gecer2017color} proposed COSFIRE filters based on colour-blob concepts, which are selective to a combination of diffused circular patches (blobs) in a particular mutual spatial agreement. These filters integrate the outputs from selected Difference-of-Gaussian filters across various scales, ensuring efficient recognition of the prototype objects. By incorporating the SURF descriptors with COSFIRE filters as the trainable parameter, Azzopardi et al.~\cite{azzopardi2018fusion} introduced a novel method for automatic gender identification from faces. The approach is still reliable despite varying facial expressions, positions, and lighting conditions. Simanjuntak and colleagues~\cite{simanjuntak2020fusion} explored the fusion of convolutional neural networks (CNNs) with the COSFIRE method to recognize gender from facial images. By incorporating COSFIRE filters into a pre-existing CNN model, they achieved enhanced accuracy and reduced the error rate by over $50\%$.

Liu et al.~\cite{liu2017computational} explored a V1-MT bio-inspired model for recognizing human actions. This model mimics neural processes in two visual cortical regions: the V1 (primary visual cortex) and the MT (middle temporal cortex) associated with movement. The design incorporates layered spike neural networks (SNNs) that integrate specific neuron properties from V1 and MT. This involves creating channels for different speeds and directions, utilizing 3-D Gabor filters and 3-D differences of Gaussian functions, and then converting motion information into spike trains. These trains are transformed into a mean motion map, a new feature vector representing human actions. In another work, Doutsi et al.~\cite{doutsi2018retina} presented a novel filter that draws inspiration from the three layers of the human retina, especially the Outer Plexiform Layer (OPL). The key feature of this filter is its basis on the linear transformation in the OPL, which is mathematically described as the ``virtual retina". This transformation is characterized by spatio-temporal weighted differences of Gaussian (WDoG) filters. These filters can shift from a lowpass to a bandpass in their spectral properties, allowing the extraction of varied information from the image. Lin et al.~\cite{lin2018contour} investigated the characteristics of neurons in the human primary visual cortex (V1) that have both classical and non-classical receptive fields (CRF and nCRF). The nCRF typically inhibits responses to visual stimuli similar to those in the CRF. Using the linear and non-linear features of retinal X and Y cells as a basis, the researchers created a contour detection model. Hu et al.~\cite{hu2019figure} introduced a recurrent neural network inspired by biological processes to gain insights into the brain's functioning, especially in understanding the cortical structures responsible for figure-ground organization. Their model uses edge detection cells and grouping cells, which recognize the preliminary objects by integrating local features. The model aligns with the neurophysiological observations regarding the timing and uniformity of border ownership cues across different scenes.

Based on the center-surround contrast, Cao et al.~\cite{cao2019application} introduced a technique that derives center-surround contrast details from natural images through the use of a normalized difference of Gaussian function combined with a sigmoid activation function. This approach is adept at rapidly and precisely reducing textures, yielding better contour detection. Melotti et al.~\cite{melotti2020robust} presented a novel model that exhibits Push-Pull inhibition and surround suppression, resulting in an enhanced contour detection method. This model minimizes texture responses while emphasizing strong reactions to lines and edges. Combining the CORF model with Push-Pull inhibition and integrating surround suppression significantly enhances the performance of the model. Nicola et al.~\cite{strisciuglio2020enhanced} introduced a 'Push-Pull' layer, inspired by biological models, for Convolutional Neural Networks (CNNs). This layer enhances the network's resilience to the various corruptions to the input image.
Furthermore, they demonstrated that this layer significantly improved the classification accuracy for corrupted images while maintaining the SOTA (state-of-the-art) performance on standard image classification tasks. Using the Push-Pull inhibition, George et al.~\cite{bhole2022corf3d} proposed a non-invasive method for identifying individual Holstein cattle leaving milking stations using thermal-RGB cameras. The study uses thermal images to separate cows from the background and occluding obstructions and employs both RGB and a novel CORF-3D contour map to identify the cattle effectively. Lin et al.~\cite{lin2022bio} proposed a model influenced by biological mechanisms that employ sub-field-based inhibition. This framework integrates inhibition terms by blending center-surround and surround-surround discrepancies, considering factors like energy distribution based on orientation and the prominence of directions in different regions. The comprehensive experimentation in \cite{lin2022bio} suggests that it can outperform other methods, particularly those driven by biological principles.

\section{PUSH-PULL CORF MODEL}
We shall use the Push-Pull CORF (Combination of Receptive Fields) model of simple cells in the primary visual cortex as feature extractors for convolutional neural networks. Hence, we summarize the CORF model \cite{azzopardi2012corf} and its extension Push-Pull CORF model\cite{azzopardi2014push} to a reasonable extent so that the computational steps become clear to the readers.

\subsection{Ganglion Cell and Lateral Geniculate Nucleus}

There are different types of Retinal Ganglion Cells that receive visual stimuli, and they do a wide variety of processing, including extraction of attributes relating to color, texture, motion, and contrast~\cite{kimetal}. This information is passed to the Lateral Geniculate Nucleus (LGN) in the thalamus. There are two LGNs, one on the left side and the other on the right side of the thalamus, each having six layers of neurons.  Neurons of the LGNs have a direct pathway to the primary visual cortex. The LGN cells respond to changes in contrast in the visual scenes. Their responses are based on concentric receptive fields in the retina. Only the information that generates a response greater than the action potential threshold is transmitted to the brain. In this study, we shall not consider the processing done by the Retinal Ganglion cells, and hence, we shall not discuss it further.

The CORF  model is a model of the simple cell in the primary visual cortex essentially based on the responses of the LGN cells having center-surround receptive fields \cite{azzopardi2012corf}. 
Receptive fields are of two types:
\begin{itemize}
    \item On-center receptive field: It responds strongly to light hitting its central area. The responses decrease as we shift towards the periphery.
    \item Off-center receptive field: It responds strongly to light hitting its peripheral area. The responses diminish as we approach closer to the center.
\end{itemize}

Hubel and Wiesel \cite{hubelwiesel1962receptive}, based on their classic experiment of 1962, suggested that the direction selectivity of simple cells and the profiles of their receptive fields (RFs) are the consequence of the RFs of the LGN cells that provide input to the simple cells. The CORF model of simple cells appears to be built on this concept. The CORF model involves several sub-units. Each sub-unit is represented by a pool of model LGN cells having either center-on or center-off receptive fields. The receptive field of a sub-unit is the union of the RFs of the involved model LGNs. Conceptually, a sub-unit that gets input from a set of center-on or center-off LGN cells may be thought of as dendritic branches of a simple cell that receives inputs from a set of adjacent LGN cells. Thus, like an individual LGN cell, a sub-unit detects changes in contrast over a bigger area. The direction/orientation selectivity of a simple cell under the CORF model is found by aggregating the responses of two sets of sub-units, one with center-on (positive) and the other with center-off (negative) LGNs. The model LGN cell has been used to realize the Push-Pull CORF model cell \cite{azzopardi2014push}. This approach extends the CORF model \cite{azzopardi2012corf} cell by incorporating Push-Pull inhibition.

\begin{figure}
    \centering
    \includegraphics[width=9cm, height=3cm]{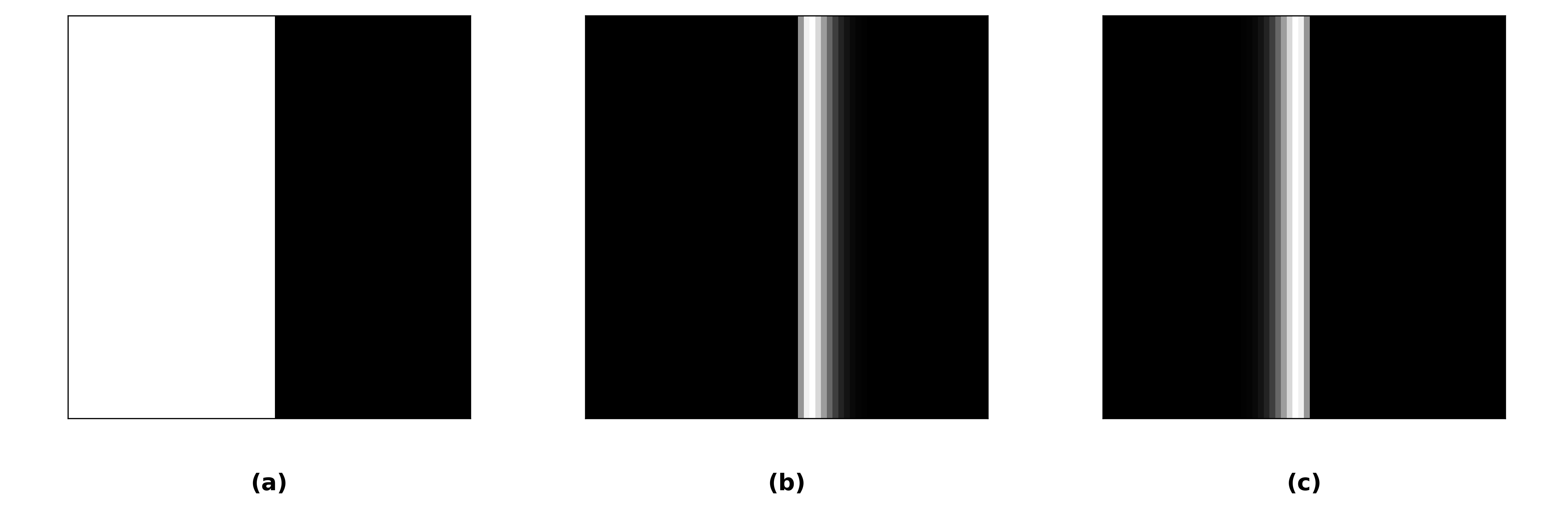}
    \caption{\textbf{(a)} Synthetic Stimulus with a vertical edge, and its output \textbf{(b)} center-on model LGN cell and \textbf{(c)} center-off model LGN cell}
    \label{fig: stimulus image}
\end{figure}

Like many other studies, to model the contrast sensitivity of LGNs, the difference of two  2-D Gaussian functions is used. In other words, the function of an LGN cell is modeled using the difference of two 2-D Gaussian functions. The CORF model realizes the direction selectivity of a simple cell by aggregating appropriate polarities and receptive fields of sub-units.  This is depicted in Fig.~\ref{fig: stimulus image}. Fig.~\ref{fig: stimulus image}(a) shows a sharp change in intensity, while Figs.~\ref{fig: stimulus image} (b) and (c) show the responses of a center-on and center-off model LGN cells, respectively. Observing the left and right sides of the cell's center, these black and white areas enable the LGN cell to detect changes in the contrast in the horizontal direction. One can manipulate the CORF model's parameters to detect contrast changes in any orientation. In the CORF model, responses to contrast changes from various directions are collected and fused to produce a comprehensive response.

As indicated earlier, the center-surround receptive field of the LGN cell is modeled using the difference between two Gaussian functions: the inner Gaussian function (G1) and the outer Gaussian function (G2). The standard deviations for these Gaussian functions are set to $\sigma$ and $\sigma/2$, respectively. Electrophysiological findings from the LGN cells in mammals influenced the selection of the standard deviation for the inner Gaussian~\cite{irvin1993center}.


\begin{equation} \label{one}
    DOG_\sigma^+(x,y) = \frac{1}{2\pi(0.5\sigma)^2}\exp{\frac{-(x^2 + y^2)}{2(0.5\sigma)^2}} - \frac{1}{2\pi\sigma^2}\exp{\frac{-(x^2 + y^2)}{2\sigma^2}}
\end{equation}

In the center-on receptive field, the central area activates the response, while the surrounding area suppresses it. This can be denoted by $DOG_\sigma^+(x,y)$ as in equation \eqref{one}. Conversely, in the center-off receptive field, its central area suppresses the response, and the surrounding area activates it. This is denoted by $-DOG_\sigma^+(x,y)$, the negative of the center-on receptive field representation, see equation \eqref{two}. 
\begin{equation} \label{two}
    DOG_\sigma^-(x,y) = - DOG_\sigma^+(x,y)
\end{equation}   

The activity of an LGN  cell is determined by combining the intensity pattern from the input image and weighing it with the DOG response as in equation \eqref{three}.
\begin{equation} \label{three}
    C_\sigma^\delta(x,y) = |I*{DoG}_\sigma^\delta|^+
\end{equation} 
         
Here, $\delta$ signifies the receptive field's polarity, the symbol $*$ stands for the convolution process, and $|.|^+$ indicates half-wave rectification. In equation \eqref{three}, the response of a receptive field centered at $(a, b)$ is determined by linearly summing up the intensity distribution $I(a, b)$, weighted by $DOG(x-a, y-b)$.

\begin{figure} 
    \centering
    \includegraphics[width=5cm, height=5cm]{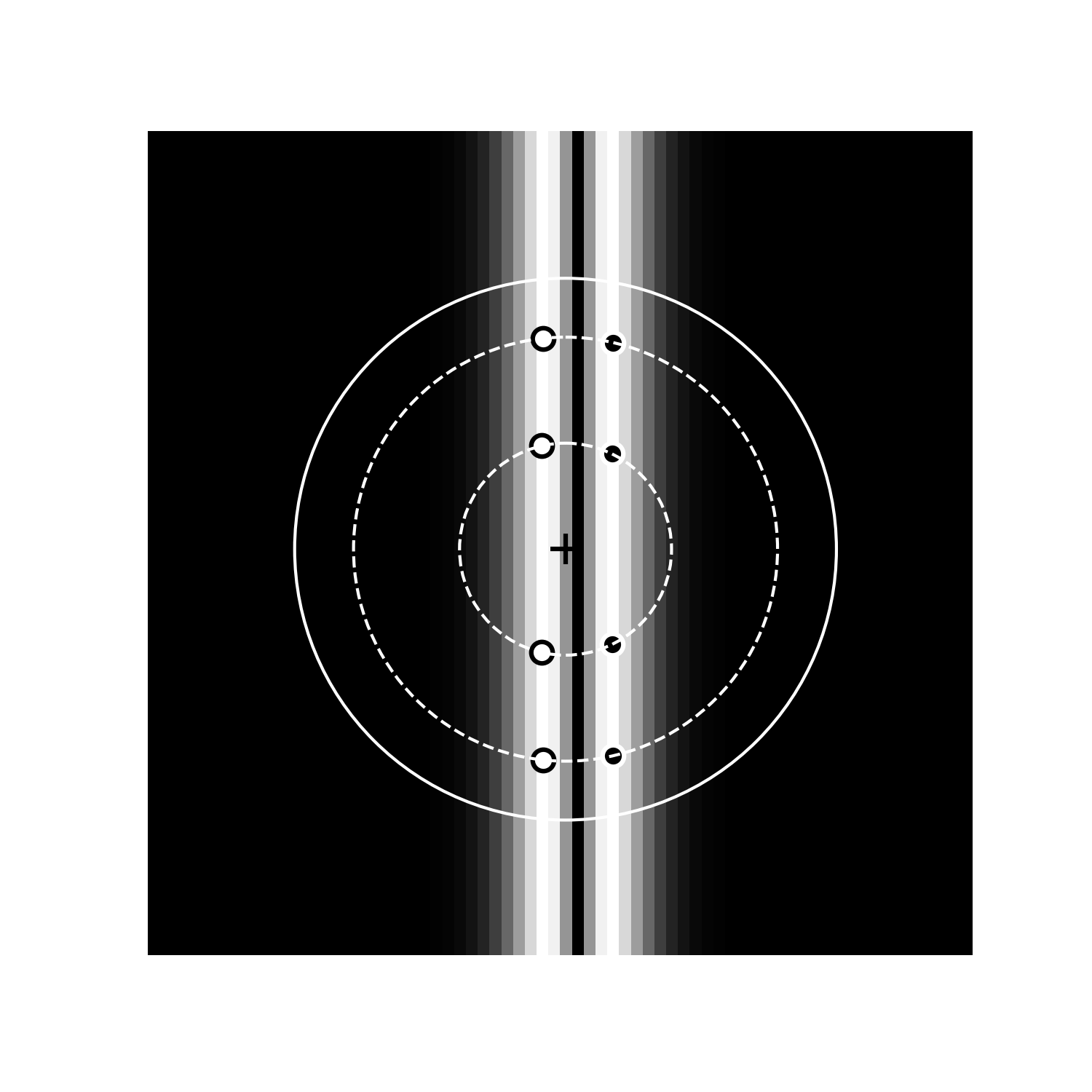}
    \caption{Setup of CORF model cell for a Stimulus with a sharp vertical edge. This center of the CORF model is marked by a ``+" on the sharp vertical edge. The larger circle denotes the receptive field boundary of the CORF Model. The marked tiny points represent the receptive field centers of eight sub-units: four are Center-on sub-units (shown as white dots with a black border), and four are Center-off sun-units (indicated by black dots with a white border)}
    \label{fig: corf model cell}
\end{figure}

 The CORF model cell is designed by superimposing the peak (local maximum) responses of both DoG+ and DoG- stimuli. We use \( k \) concentric circles centered at the center of the receptive field (RF) of the model simple cell, where \( k \) is determined by \( \sigma \). We identify the peak values (local maxima) along these circles. A circular area around each peak represents the RF of a sub-unit (Fig. \ref{fig: corf model cell}). Each circle has four sub-units: two center-on sub-units (white dot against a black background) resulting from the DoG+ stimulus response and two center-off sub-units (black dot against a white background) obtained from the DoG- stimulus response using equation \eqref{three}.

Azzopardi et al.\cite{azzopardi2012corf} implemented a hyperparameter-based method to determine radii based on the $\sigma$ value. Depending on the  $\sigma$ values, concentric circles of varying radii are considered. A four-tuple characterizes the properties of a pool of afferent  LGN model cells as $s = (\delta, \sigma, \rho, \phi)$, where    $\delta$ represents the polarity of the sub-unit (-1 for center-off and +1 for center-on), $\sigma$ is the spread of the Gaussian that is utilized to model the LGN cell, the $\rho$ indicates the radius and $\phi$ denotes the polar angle about the center of the receptive field of the sub-unit with respect to the receptive field center of the CORF model cell. Define $S = \{(\delta_i, \sigma_i, \rho_i, \phi_i) | i = 1, \cdots, n \}$; in Fig. \ref{fig: corf model cell}; $n=8$. The S represents the set of sub-units corresponding to the preferred orientation (in Fig. \ref{fig: corf model cell}, it is the vertical direction) of the simple cell modeled by the sub-units. Next, we calculate the response of each sub-unit, which will be combined to compute the response of a simple cell.

The response of a sub-unit is calculated by linearly combining the half-wave rectified outputs of the LGN model cells. This computation considers the CORF model cell's receptive field center, considering the polarity denoted as $\delta$, the scaling factor $\sigma$, and centered at position $(\rho, \phi)$. This is further modulated by a Gaussian weight, represented as $G_{\sigma}.\odot$.

\begin{equation} \label{four}
    s_{{\delta},{\sigma},{\rho},{\phi}}(x,y) = \sum_{a'}\sum_{b'} {c_\sigma^\delta(x - \Delta a - a', y- \Delta b - b')}G_{\sigma'} (a', b')
\end{equation}

Here, $\Delta a = -\rho \cos{\phi}$ and $\Delta b = -\rho \sin{\phi}$. The values of $a'$ and $b'$ are inclusively within the range of $-3\sigma'$ to $3\sigma'$. The standard deviation, represented as $\sigma'$ is directly related to the $\rho$ and is influenced by some other parameters~\cite{azzopardi2012corf}. 
The output $c_\sigma^\delta(x,y)$ is adjusted by $\Delta a$ and $\Delta b$ based on the value of $\rho$ and $\phi$ at location $(a', b')$.

Finally, the output of the CORF simple cell is determined by taking the weighted geometric mean of the responses from all sub-units within the set S as in equation \eqref{five}. If the distance of a sub-unit from the CORF model cell's center increases, then its weight diminishes.


\begin{equation}\label{five}
    {r_S (x,y)} = {\prod_{i=1}^n (s_{\sigma_i, \delta_i,{\rho},{\phi}} (x,y))^{w_i})}^{(\frac{1}{\sum_{i=1}^{n} w_i})}
\end{equation}\\

where $w_i$ is a Gaussian weight defined by $\rho_i$ and $\sigma_i$ \cite{azzopardi2012corf}. 
A CORF model cell responds strongly to the configured orientation when all the sub-units are active. It declines as the stimulus orientation moves away from the optimal one, and it almost becomes negligible for deviation larger than $\frac{\pi}{4}$. To calculate the responses along different orientations, we can manipulate the polar angle $\phi$ as $\Re_\psi(S)=\{(\delta_i, \sigma_i, \rho_i, \phi_i+\psi)|\forall(\delta_i, \sigma_i, \rho_i, \phi_i) \in S  \}$. The computational steps are summarized in Algorithm-\ref{alg:corf_response}. Authors in \cite{azzopardi2012corf} have demonstrated the success of the CORF model in extracting edges on some well-known datasets. However, the CORF model cell fails to realize some of the characteristics of the actual (biological)  simple cells, such as the contrast-dependent changes/selectivity in spatial frequency. To account for such properties of the actual simple cells, the CORF model is extended to the Push-Pull CORF model in \cite{azzopardi2014push}.

\subsection{The Push-Pull Extension}
There exists neuro-science evidence that some simple cells exhibit Push-Pull inhibition. In \cite{azzopardi2014push}, the CORF model is extended to account for the Push-Pull inhibition. In equation \eqref{five}, the response of a CORF model cell for a stimulus of preferred orientation and contrast at $(x, y)$ is computed as $r_s(x, y)$. Let $r_{\hat{S_{\beta}}}(x,y)$ be the response of a CORF model cell for the same stimulus but of opposite contrast. Then the response of a Push-Pull CORF model cell at $(x,y)$ is computed as $ r_{P_{\beta}}=r_{S}(x,y) - k*r_{\hat{S_{\beta}}}(x,y)$. A separation measure is used to model the Pull inhibition \cite{azzopardi2014push}. The separation is the distance between a pool of On-center and Off-center LGN cells. In the receptive field arising from the CORF model cell (as shown in Fig.~\ref{fig: corf model cell}), the excitatory and inhibitory areas may be separated in the direction perpendicular to the preferred orientation. This separation is the distance between a set of center-on and center-off model LGN cells. Let the distance be $B = \epsilon$. Setting this separation  as $B = \epsilon + \beta$, authors in \cite{azzopardi2014push} created a new set $S_{\beta}$, representing another simple CORF model cell for the same preferred orientation as follows:

\begin{equation} \label{six}
S_{\beta} = \{(\delta_i, \sigma_i, \rho_{i}^{'}, \phi_{i}^{'})|\forall(\delta_i, \sigma_i, \rho_i, \phi_i)\in S\}
\end{equation}

where 
$\rho_{i}^{'}=\sqrt{(x_i+\gamma)^2+y_i^2}$, 
$\phi_i=\arccos(\frac{x_i+\gamma}{\rho_{i}^{'}})$, 
$x_i=\rho_{i}\cos\phi_{i}$, $y_i=\rho_{i}\sin\phi_{i}$, 
$\gamma=\beta/2$ 
when $x_{i}>0$ and $\gamma=-\beta/2$ when $x_{i}<0$.

 The magnitude of parameter $B$ influences the intensity of response to the stimulus. It also impacts the orientation bandwidth and spatial frequency of the CORF model cell. As the value of $\beta$ increases, the orientation bandwidth increases while the spatial frequency and the response to the stimuli decreases. To obtain the response of the simple cell for the same stimulus with the same preferred orientation but opposite contrast, the updated CORF model cell is defined as \cite{azzopardi2014push}:
\begin{equation} \label{seven}
\hat{S_{\beta}} = \{(-\delta_i, \sigma_i, \rho_{i}^{'}, \phi_{i}^{'})|\forall(\delta_i, \sigma_i, \rho_i, \phi_i)\in S\}
\end{equation}

Using this Pull inhibition, the response of the Push-Pull CORF model cell at $(x,y)$ is defined as 
        \begin{equation} \label{eight}
             r_{P_{\beta}} = r_{S}(x,y) - k*r_{\hat{S_{\beta}}}(x,y)
        \end{equation}

In equation \eqref{eight}, $r_{S}$ and $r_{\hat{S}}$ denote the Push response and Pull response, respectively, and the factor $k$ represents the strength of the Pull inhibition. 

\begin{figure}
    \centering
    \includegraphics[width=\textwidth]{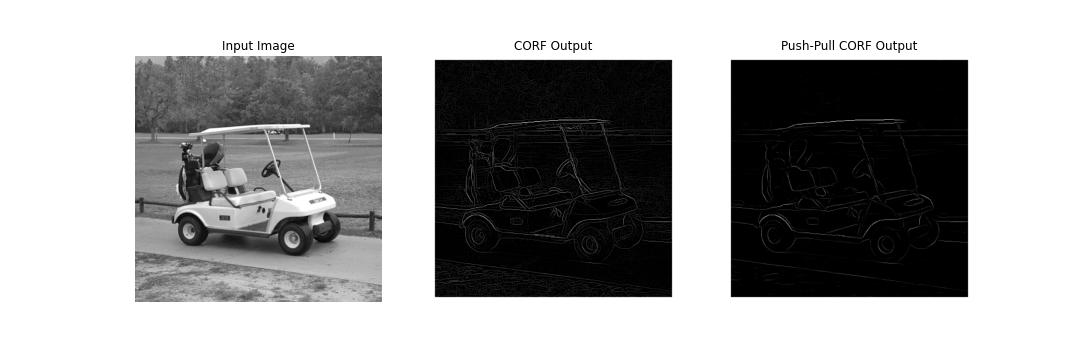}
    \caption{Comparison of the Response generated by CORF model and Push-Pull CORF model }
    \label{fig: output_comparison}
\end{figure}

\section{Network Architecture}

\begin{figure}
    \centering
    \includegraphics[width=0.5\textwidth]{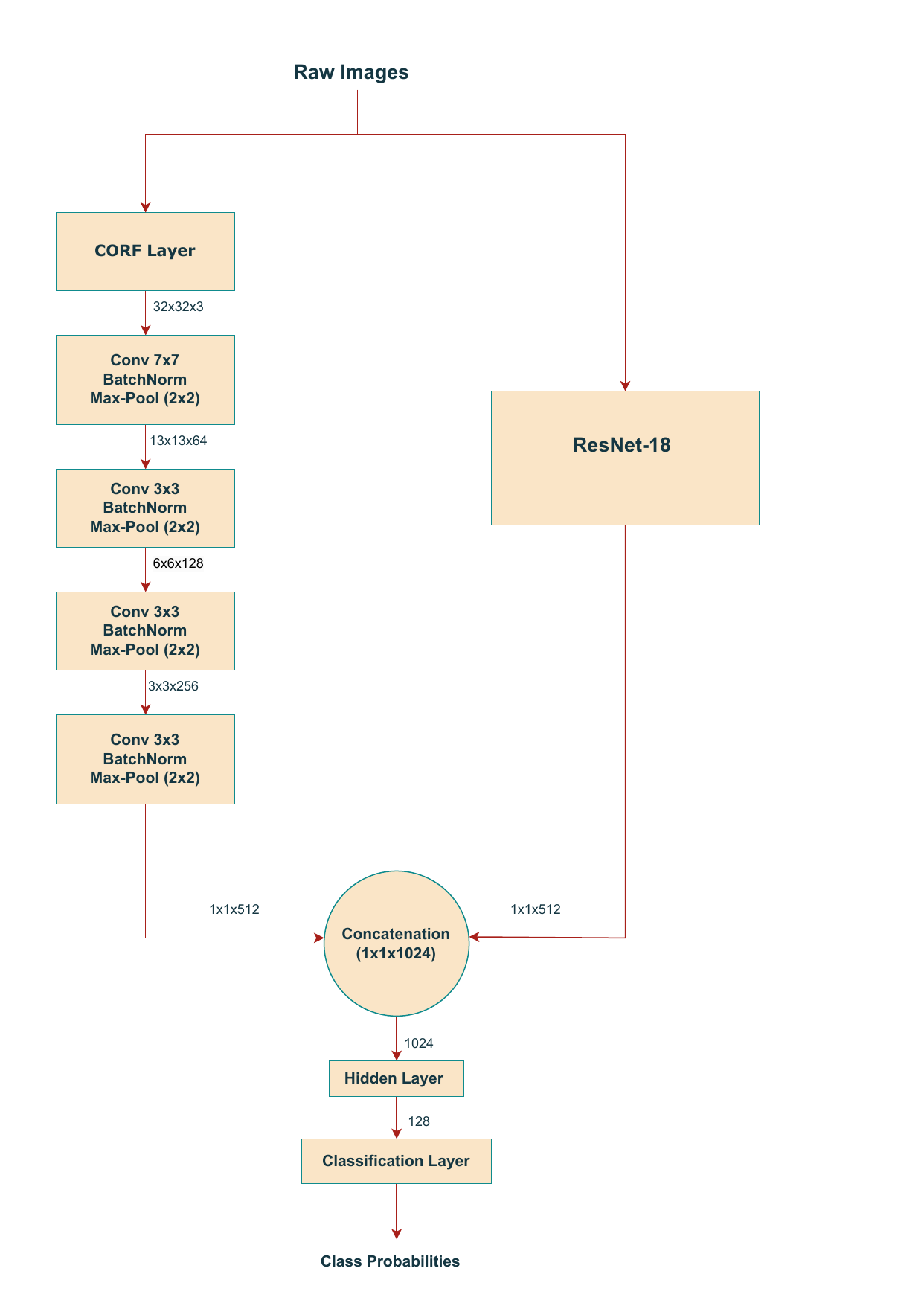}
    \caption{Two-Tower Architecture: the first tower is a shallow network containing a Push-Pull CORF layer, and the rest of the four blocks are convolution blocks with a max-pooling layer. The second tower is a ResNet-18 architecture. Feature Maps from both the towers are fused using concatenation, which is passed through the full-connected layers for classification task }
    \label{fig:two-tower architecture}
\end{figure}

Our architecture (shown in Fig.~\ref{fig:two-tower architecture}) is composed of two distinct towers:
\begin{itemize}
    \item A shallow network with a Push-Pull CORF model simple cells as its initial layer. This layer is intended to extract some of the features that the simple cells extract in conjunction with the LGN cells.
    \item The ResNet-18 architecture
\end{itemize}

\subsection{First Tower}
This tower is comprised of five layers. Its initial layer is the Push-Pull CORF layer, succeeded by four 2D convolution layers. The CORF layer processes raw images and computes the responses of both the center-on DOG operator and the center-off DOG operator. 

For a given $\sigma$ and orientation, we configured the CORF model cell. Depending on the $\sigma$, a number of concentric circles are considered centered at the center of the image. The local maxima locations are identified on each circle, where each maxima corresponds to a sub-unit of a simple cell. The half-wave rectified response of this LGN cell was linearly combined to determine sub-unit responses weighted by a Gaussian function. The Push response of the CORF model cell was ascertained by combining these sub-unit responses, adjusted using parameters $\delta$, $\rho$, and $\phi$, leading to the computation of the response of a simple CORF model cell for stimuli of preferred orientation with varied contrast. This process produced the response of the Push CORF model. For each $\sigma$ value, we have calculated 12 output images (in 12 orientations from $0^{\circ}$ to $360^{\circ}$, at an interval of $30^{\circ}$). A single output response is calculated from these 12 images by taking pixel-wise maximum superposition.
Similarly, we computed the inhibitory Pull CORF model response. Then, combining the Push CORF and Pull CORF responses, we get a single image. Utilizing 17 different $\sigma$ values, we obtained 17 unique output responses, which form a 17-channel input for the subsequent layers. In the Push-Pull CORF layer, the initial $\sigma$ values, vary ranging from 1 to 5 at 0.25 intervals. The sub-unit locations are determined using these initial values and remain unchanged during training. However, the $\sigma$ utilized in computing the sub-unit response is subject to updates during the training phase.

After passing through the Push-Pull CORF Layer, the data is directed through multiple convolution layers. Every one of these layers is equipped with batch normalization and max-pooling capabilities. The layers in Tower 1 are detailed as follows:
\begin{itemize}
    \item a Push-Pull CORF Layer with $\sigma$ values from 1 to 5 at intervals of 0.5 and an inhibition factor $k=1.8$
    \item a convolution layer with 64 filters, each of size 7x7, with max-pooling of size 2x2
    \item a convolution layer equipped with 128 filters of size 3x3, with max-pooling of size 2x2
    \item a convolution layer with 256 filters of size 3x3, with max-pooling of size 2x2
    \item a final convolution layer with 512 filters of size 3x3, with max-pooling of size 2x2
\end{itemize}

\subsection{Second Tower}
The second tower employs the ResNet-18 architecture, which belongs to the ResNet (Residual Network) family. This structure is engineered to facilitate the training of profoundly deep neural networks by capitalizing on residual (or skip) connections. The standout feature of ResNet is the "residual blocks" that allow for the direct relay of input to the network's deeper layers. This strategy addresses the vanishing gradient issue commonly encountered in deep networks. The layers of ResNet-18 are structured as follows:
\begin{itemize}
    \item an initial convolutional layer with 64 filters of size 7x7 and stride of 2.
    \item a max pooling of size 3x3 with a stride of 2.
    \item two blocks of 2 layers, each with 64 filters of size 3x3.
    \item two blocks of 2 layers, each with 128 filters of size 3x3.
    \item two blocks of 2 layers, each with 256 filters of size 3x3.
    \item two blocks of 2 layers, each with 512 filters of size 3x3.
    \item an adaptive average pooling layer.
\end{itemize}

In every block, there are two convolutional layers. The input to the block is combined with its output using a residual connection. If the number of channels changes in the block, then the dimensions are matched using a 1x1 convolution in the shortcut path. After each convolution operation, batch-normalization and ReLU are used. ResNet employs global average pooling to reduce the spatial dimensions to 1x1. This ensures a fixed output size, which is important for handling images of varying sizes. Additionally, the technique improves the model's ability to handle spatial variations by focusing on the presence of features rather than their specific locations.

\subsection{Fusion of Feature Maps}
Once the feature maps from both towers are generated, they are fused using a concatenation operation. This merged feature map then navigates through a sequence of two fully connected layers: (a) a hidden layer comprised of 128 neurons and (b) a layer dedicated to classification.

\begin{algorithm*}
\caption{CORF Response Calculation}\label{alg:corf_response}
\renewcommand{\algorithmicrequire}{\textbf{Input:}}
\renewcommand{\algorithmicensure}{\textbf{Output:}}
\begin{algorithmic}
\Require An input image with a sigma ($\sigma$) value
\Ensure A CORF output image

\Procedure{CORF}{$\sigma$}
\For{a given $\sigma$ and orientation}
    \State {At each pixel $(x,y)$, compute the response of a model LGN cell (Center-On RF) with RF centered at $(x,y)$ using \eqref{three} with DoG+}
    \State Draw circles centered at the center of the image (number of circles depends on the $\sigma$)
    \State Find local strong maxima on each circle
    \State Each maxima corresponds to a sub-unit of a simple cell
    \State {Each sub-unit collects responses from the LGN cells with polarity $\delta$ within the receptive field defined by $\sigma$ (say 5 pixels) around the position defined by ($\rho$, $\phi$) with respect to the RF-center of the CORF model cell. The sub-unit response is computed using the equation\eqref{four}}
    \State Repeat the above steps using Center-Off RF with DoG-
\EndFor

\State Combine responses of all sub-units using weighted geometric mean (AND operation) to get the CORF image for given $\sigma$ and orientation
\State Use 12 different orientations (biological evidence) and get 12 different CORF images
\State Combine these 12 images using pixel-wise maximum superposition operation $\rightarrow$ Final CORF image for a given $\sigma$

\EndProcedure
\end{algorithmic}
\end{algorithm*}


\section{Experimental SetUp}
\subsection{Datasets}
We utilize three datasets for image classification: CIFAR-10~\cite{krizhevsky2010cifar}, CIFAR-100~\cite{krizhevsky2012cifar}, and ImageNet-100, which is a selection of the random 100 classes from the original ImageNet~\cite{imagenet} dataset. Our approach incorporates data augmentation methods, including random flips, cropping, and resizing.

\subsubsection{CIFAR-10}
It contains 60,000 RGB images with dimensions of 32x32. These images are categorized into ten classes- aeroplanes, birds, cars, cats, deer, dogs, frogs, horses, ships, and trucks. The dataset is divided into training and testing sets with 50,000 and 10,000 images. 

\subsubsection{CIFAR-100}
The CIFAR-100 dataset, originating from the Canadian Institute for Advanced Research and featuring 100 categories, is a part of the Tiny Images collection. It contains 60,000 colour images, each with a resolution of 32x32 pixels. These 100 categories are organized under 20 broader categories called superclasses. Every individual class has 600 images. Each image is labelled with its specific class ("fine" label) and its overarching category ("coarse" label). For each class, 500 images are designated for training and 100 for testing. 

\subsubsection{ImageNet-100}
As mentioned earlier, the ImageNet-100 is a smaller selection from the ImageNet-1k Dataset, originating from the 2012 ImageNet Large Scale Visual Recognition Challenge. This subset considers 100 classes chosen at random from the original 1,000 classes. The training set has 1,300 images for each class, while the test set has 50 images per class.

\subsection{Hyper-Parameter Setting}
We used five different random seeds for our experiment. We utilize a batch size of 128 in every experiment for each dataset. The Stochastic Gradient Descent~\cite{amari1993backpropagation} optimizer is adopted with an initial learning rate of 0.2, a momentum of 0.9, and the weight decay parameter set at \num{5e-4}. The learning rate scheduler is based on Cosine Annealing~\cite{loshchilov2017decoupled}. We used Cross-Entropy loss as the loss function. The training spans 2000 cycles, integrating an early-stopping~\cite{yao2007early} mechanism with a patience of 200 epochs. The best-performing model on validation loss is stored. After every epoch, the loss achieved up to that point is assessed. The model weights are saved if the current validation loss is superior to the previous best.

\subsection{Performance Evaluation Measures}
To assess the performance of our proposed model with ResNet-18 architecture, we use four popular metrics- Accuracy, Precision, Recall, and Macro-F1 score.
\begin{itemize}
    \item Accuracy represents the proportion of correctly identified instances out of all instances. It is given by: 
    $$accuracy = \frac{TP + TN}{TP + FP + TN + FN}$$
    Here, TP, TN, FP and FN stand for true positive, true negative, false positive, and false negative, respectively.
    \item Precision indicates the fraction of accurately predicted positive instances out of the total predicted positive instances.
    $$precision = \frac{TP}{TP + FP}$$
    \item Recall (sensitivity) measures the proportion of actual positives that are correctly predicted.
    $$recall = \frac{TP}{TP + FN}$$
    \item The F1 Score is the weighted average (harmonic mean) of precision and recall. It is calculated as:
    $$F1-Score = \frac{2 \times precision \times recall}{precision + recall}$$
\end{itemize}

\subsection{Experimental Results}
For our study, we divided the training set of each dataset into a train set and a validation set using an 80:20 ratio. A comparison of our proposed architecture with ResNet-18 architecture is shown in Table~\ref{tab: result_ppcorf}. As mentioned earlier, we use three commonly used datasets for image classification tasks. Our architecture outperforms the ResNet-18 model, showing a Macro F1-Score improvement of 5\%-10\%. Specifically, we achieve a $5\%$ increase on the CIFAR-10 dataset, a $10\%$ increase on the CIFAR-100 dataset, and an $8\%$ increase on the ImageNet-100 dataset. The accuracy also increased from $5\%$ to $11\%$.

\begin{figure}
    \centering
    \includegraphics[width=\textwidth]{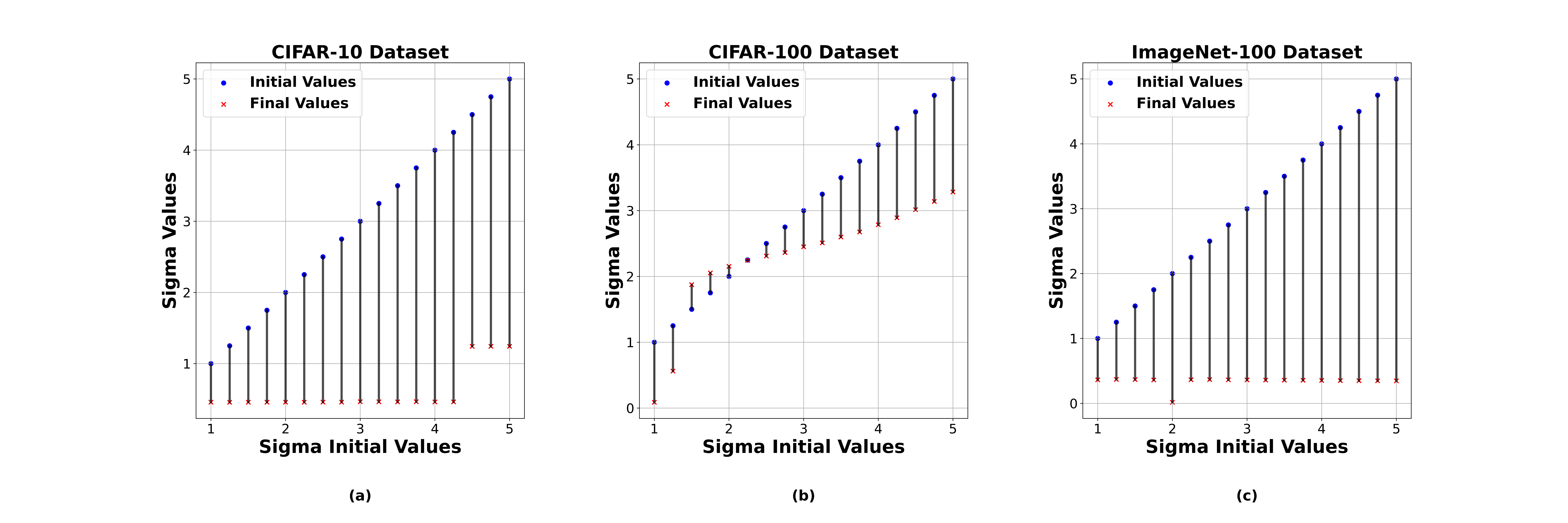} 
    \caption{Transition of Sigma values: (a) in the CIFAR10 dataset, it is optimizing to two different points- 0.45 and 1.24, (b) for the CIFAR100 dataset, it is optimizing to a range of values from 0.087 to 3.283, and (c) for ImageNet-100, it is optimizing to 0.36.
    }
    \label{fig: Transition of Sigma values in training}
\end{figure}

Fig.~\ref{fig: Transition of Sigma values in training} depicts the updated sigma ($\sigma$) values for the three datasets. Initially, the sigma values are from 1 to 5 at an interval of 0.25. For the CIFAR-10 dataset, the $\sigma$ values are optimized to two points, $0.46$ and $1.24$. For the CIFAR-100 dataset, the $\sigma$ values are optimized from 0.0877 to 3.2826, while for the ImageNet-100 dataset, the $\sigma$ values are optimized to 0.35 and 0.36, except for one $\sigma$ value 2, which is optimized to 0.0164. The inhibition factor ($k$) is initially set to $1.8$, which is optimised to $1.37$, $1.64$, and $1.13$ for CIFAR-10, CIFAR-100, and ImageNet-100, respectively.


\begin{table*}[ht]
    \centering
    \small 
    \caption{Result of proposed architecture and ResNet-18 architecture on CIFAR-10, CIFAR-100, and ImageNet-100}
    \begin{tabularx}{\textwidth}{|c|X|X|X|X|X|X|X|X|}
        \hline
        \multirow{2}{*}{\textbf{Datasets}} & \multicolumn{4}{c|}{\textbf{ResNet-18}} & \multicolumn{4}{c|}{\textbf{Our Proposed Model}} \\
        \cline{2-9}
        & \centering Accuracy & \centering Precision & \centering Recall & \centering\arraybackslash Macro-F1 & \centering Accuracy & \centering Precision & \centering Recall & \centering\arraybackslash Macro-F1 \\
        \hline
        CIFAR-10 & ${0.86\pm0.02}$ & ${0.86\pm0.02}$ & ${0.86\pm0.02}$ & ${0.86\pm0.02}$ & ${0.91\pm0.01}$ & ${0.91\pm0.01}$ & ${0.90\pm0.01}$ & ${0.91\pm0.01}$ \\
        \hline
        CIFAR-100 & ${0.50\pm0.03}$ & ${0.50\pm0.03}$ & ${0.50\pm0.03}$ & ${0.50\pm0.03}$ & ${0.61\pm0.02}$ & ${0.61\pm0.02}$ & ${0.61\pm0.02}$ & ${0.61\pm0.02}$ \\
        \hline
        ImageNet-100 & ${0.65\pm0.03}$ & ${0.65\pm0.02}$ & ${0.65\pm0.02}$ & ${0.64\pm0.02}$ & ${0.73\pm0.01}$ & ${0.72\pm0.01}$ & ${0.73\pm0.01}$ & ${0.72\pm0.01}$ \\
        \hline
    \end{tabularx}
    \label{tab: result_ppcorf}
\end{table*}

\begin{figure}
    \centering
    \includegraphics[width=\textwidth]{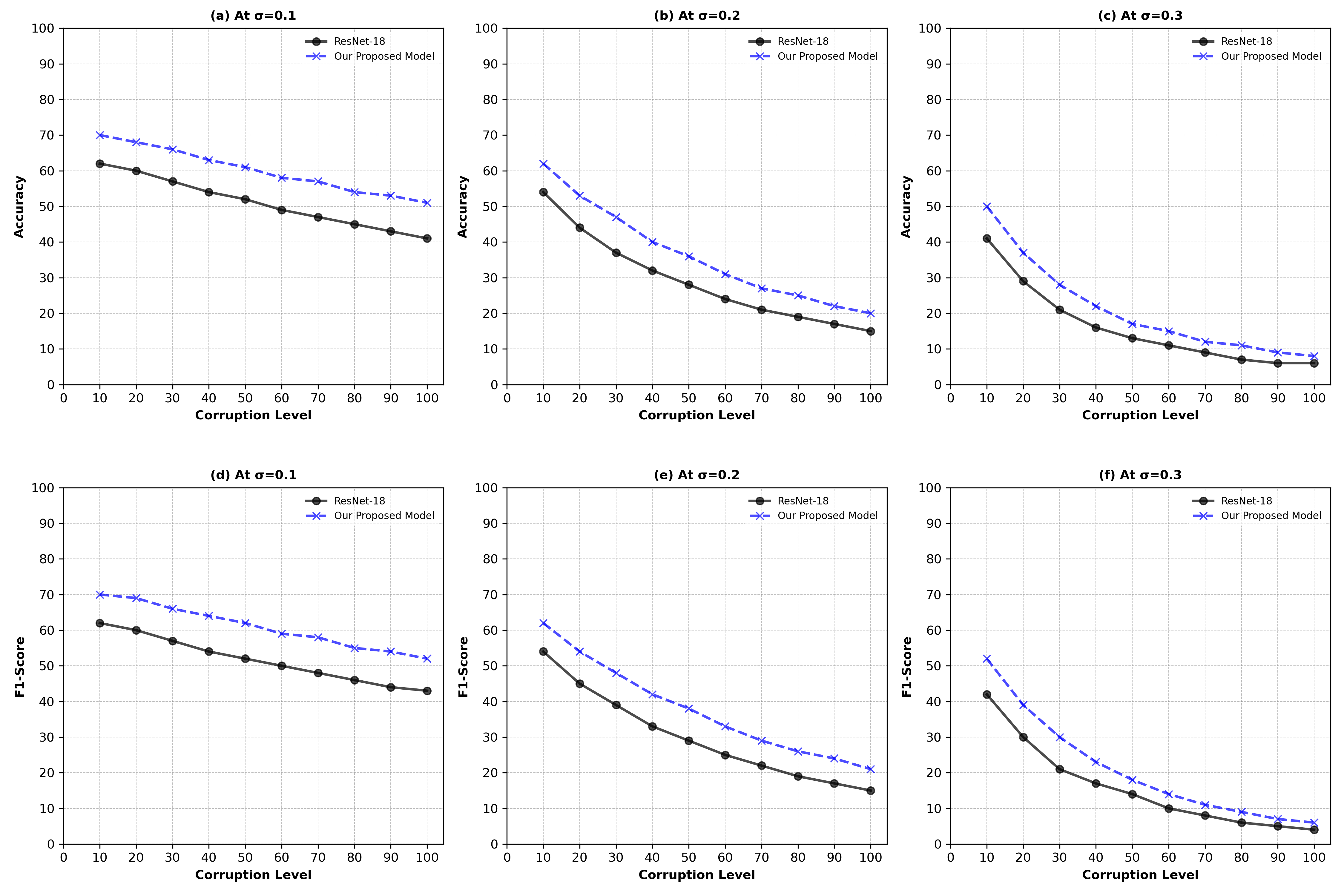}
    \caption{Performance of ResNet-18 and Our Proposed Model on ImageNet-100. Measuring Accuracy for: (a) $\sigma=0.1$, (b) $\sigma=0.2$, (c) $\sigma=0.3$, and measuring F1-Score for: (d) $\sigma=0.1$, (e) $\sigma=0.2$, and (f) $\sigma=0.3$.}
    \label{fig: noise_attack_imagenet_100}
\end{figure}

Next, we study the robustness of our model to additive Gaussian noise with different $\sigma$ and different levels of corruption. For this, we have added the Gaussian noise to pixels with percentages varying from $10\%$ to $100\%$. Fig.~\ref{fig: noise_attack_imagenet_100} has two rows. The top row compares the performance of the proposed network with that of ResNet-18 with respect to accuracies for different levels of corruption and different $\sigma$. The second row compares the F1-score with respect to different levels of corruption and different $\sigma$. For both experiments, we have used the ImageNet-100 dataset. Fig.~\ref{fig: noise_attack_imagenet_100} reveals that our model is consistently more resilient to Gaussian Noise than ResNet-18. In both accuracy and F1-score metrics, ResNet-18 demonstrated a more significant decline than our model.


The number of trainable parameters in the CORF tower is approximately $1.6$ million, while the number of trainable parameters in the ResNet-18 tower is around $11.2$ million. The total number of trainable parameters in the Two-tower architecture is approximately $13.45$ million. The number of trainable parameters in the CORF tower includes the trainable parameters of the Push-Pull CORF layer and the four-layer stack of the convolutional layers.

We have also conducted some tests to evaluate the performance of our model after removing the ResNet tower and only using the CORF tower, as shown in Table~\ref{tab: result_only_ppcorf_tower}. These results demonstrate that the model's performance decreased significantly, particularly for ImageNet-100, after dropping the ResNet tower. For the CIFAR-10 dataset, the $\sigma$ values are optimized to two values, 0.5 and 1.3. For the CIFAR-100 dataset, the $\sigma$ values are optimized to values ranging from 0.66 to 3.1. For the ImageNet-100 dataset, the $\sigma$ values are optimized to two values, 0.64 and 1.55. The inhibition factor (k) is initially set to 1.8 and optimised to 1.52, 1.68, and 1.35 for CIFAR-10, CIFAR-100, and ImageNet-100, respectively.

We also evaluated the effect of increasing the number of convolutional layers on the model's performance by increasing it from 4 to 6 (increasing the number of free parameters to 6.4 million). The results demonstrate that the performance improved significantly for the CIFAR-100 and ImageNet-100 datasets, while it remained almost the same for the CIFAR-10 dataset. For the CIFAR-10 dataset, the $\sigma$ values are optimized to 0.4 and 0.5. For the CIFAR-100 dataset, the $\sigma$ values are optimized to 0.7, 1.7, 1.8, 1.8, 1.9, 2.0 and 2.1. For the ImageNet-100 dataset, the $\sigma$ values are optimized to two values, 0.63 and 1.5. The inhibition factor (k) is optimised to 1.55, 1.71, and 1.49 for CIFAR-10, CIFAR-100, and ImageNet-100, respectively.

\begin{table*}[ht]
    \centering
    \small 
    \caption{Result of Push-Pull CORF Tower architecture on CIFAR-10, CIFAR-100, and ImageNet-100}
    \begin{tabularx}{\textwidth}{|c|X|X|X|X|X|X|X|X|}
        \hline
        \multirow{2}{*}{\textbf{Datasets}} & \multicolumn{4}{c|}{\textbf{CORF Tower (4 CNN layers)}} & \multicolumn{4}{c|}{\textbf{CORF Tower (6 CNN layers)}} \\
        \cline{2-9}
        & \centering Accuracy & \centering Precision & \centering Recall & \centering\arraybackslash Macro-F1 & \centering Accuracy & \centering Precision & \centering Recall & \centering\arraybackslash Macro-F1 \\
        \hline
        CIFAR-10 &  ${0.80\pm0.01}$ & ${0.81\pm0.01}$ & ${0.80\pm0.01}$ & ${0.80\pm0.01}$ &
            ${0.82\pm0.01}$ & ${0.82\pm0.02}$ & ${0.82\pm0.01}$ & ${0.82\pm0.02}$ \\
        \hline
        CIFAR-100 & ${0.47\pm0.02}$ & ${0.51\pm0.02}$ & ${0.47\pm0.02}$ & ${0.47\pm0.02}$ &
            ${0.51\pm0.03}$ & ${0.55\pm0.03}$ & ${0.51\pm0.02}$ & ${0.51\pm0.02}$ \\
        \hline
        ImageNet-100 & ${0.41\pm0.02}$ & ${0.53\pm0.02}$ & ${0.41\pm0.02}$ & ${0.40\pm0.02}$ &
            ${0.53\pm0.01}$ & ${0.63\pm0.01}$ & ${0.53\pm0.01}$ & ${0.53\pm0.01}$ \\
        \hline
    \end{tabularx}
    \label{tab: result_only_ppcorf_tower}
\end{table*}

\section{Conclusion}
We proposed an architecture that incorporates the feature extraction knowledge of simple cells of the primary visual cortex (which are defined using the behaviour of LGN cells) with CNNs, enabling a shallow network to improve the performance of the deep network. We aimed to extract the features like the way biological neurons do. In this context, we have used the Push-Pull CORF computational model. For the deep network, we have used ResNet-18 architecture. We first developed a basic CNN with the CORF computation as the initial layer in this setup. Our end goal was to use a two-tower architecture, one using the Push-Pull CORF-based CNN architecture and the other utilizing any deep learning architecture (here, we used ResNet-18) to boost the performance of the combined model. The outputs of the two towers are utilised by fusing the features obtained from both towers.

Our study revealed that the feature extraction incorporating characteristics of biological neurons in the shallow network significantly improves the model performance of the deep network under the same hyperparameters setting. This can be attributed possibly to some complementary features provided by the Push-Pull CORF layer. This behaviour of the model performance is expected with any other deep networks. We also observed that the proposed architecture optimizes faster than the deep network and is more resilient to additive Gaussian Noise. Furthermore, the learnable parameters of the Push-Pull CORF layer are optimized from a diverse set of values for the given hyperparameter setting.

\bibliographystyle{plain}
\bibliography{pp_corf_cnn}

\end{document}